\documentclass[conference]{IEEEtran}
\IEEEoverridecommandlockouts
\usepackage{cite}
\usepackage{amsmath,amssymb,amsfonts}
\usepackage{algorithmic}
\usepackage{graphicx}

\usepackage{threeparttable}
\usepackage{textcomp}
\usepackage{xcolor}
\usepackage{multirow}
\usepackage[table,xcdraw]{xcolor}
\usepackage{enumitem}
\usepackage{graphicx}
\usepackage{booktabs}
\usepackage{tabularx}
\usepackage{array}

\usepackage{xurl}
\usepackage[hidelinks,breaklinks=true]{hyperref}
\newcolumntype{L}{>{\raggedright\arraybackslash}X}
\newcolumntype{P}[1]{>{\raggedright\arraybackslash}p{#1}}
\usepackage{subcaption}
\usepackage{float} 
\usepackage{booktabs}
\usepackage{pgfplots}
\pgfplotsset{compat=1.18}
\def\BibTeX{{\rm B\kern-.05em{\sc i\kern-.025em b}\kern-.08em
    T\kern-.1667em\lower.7ex\hbox{E}\kern-.125emX}}
\begin{document}

\title{Agentic AI with Structured CoT for Enhancing AI's Spatial Intelligence: Visualization and Reasoning of Rotation \\

}

\author{\IEEEauthorblockN{Uttamasha Monjoree}
\IEEEauthorblockA{\textit{Department of Architecture} \\
\textit{Texas A\&M University}\\
College Station, USA \\
uxm190002@tamu.edu}
\and

\IEEEauthorblockN{Wei Yan}
\IEEEauthorblockA{\textit{Department of Architecture} \\
\textit{Texas A\&M University}\\
College Station, USA \\
wyan@tamu.edu}
}
\maketitle

\begin{abstract}
Spatial intelligence has a significant impact on many fields, such as Science, Technology, Engineering, and Mathematics (STEM), Architecture, Engineering, and Construction (AEC), Medicine, and Manufacturing. Recent studies show that artificial intelligence (AI) with language and vision capabilities still experiences limitations in spatial reasoning. In this paper, we have studied the spatial capabilities of advanced generative AI to understand the rotations of objects in 3D space, utilizing AI's image processing and language processing features. We trained and examined the spatial intelligence of a generative Agentic AI model (GPT-5.6) to understand the spatial rotation process with rotation diagrams based on the revised Purdue Spatial Visualization Test: Visualization of Rotations (Revised PSVT:R). We improvised the Revised PSVT:R by superimposing additional graphical and contextual features to evaluate how different Chain-of-Thought (CoT) reasoning strategies influence model performance. The results indicate that structured CoT reasoning improves the spatial reasoning performance of the base GPT-5.6 model in both datasets (PSVT:R and PSVT:R with coordinate system). We used three CoT approaches - (1) Structured CoT, (2) few-shot Structured CoT, and Structured CoT with Self-optimized Prompt. The three CoT approaches evaluated in this study showed no significant performance difference. Results showed that combining structured CoT reasoning with relevant contextual information leads to considerable improvements in VLM performance on 3D rotation tasks, demonstrating the potential of agentic AI for more effective spatial reasoning. However, when contextual information is removed, structured CoT reasoning alone provides limited improvement, and the models continue to exhibit notable difficulties in understanding spatial transformations. These findings suggest that effective spatial reasoning in VLMs relies on the integration of visual, textual, and reasoning-based information in future agentic AI systems for spatial intelligence.
\end{abstract}

\begin{IEEEkeywords}
Generative AI, Spatial Intelligence, Agentic AI, Chain-of-Thought, CoT, PSVT:R
\end{IEEEkeywords}

\section{Introduction}
Spatial visualization is a significant part of spatial skills. Spatial visualization skills are described as the ability to mentally rotate, manipulate, twist, or invert 3D objects \cite{ben1988effect}, \cite{fleisig2004improving}. Spatial skill and spatial guidance can greatly benefit understanding spatial transformations and additionally support processes like assembly, construction, and manufacturing, as well as learning in AEC (Architecture/Engineering/Construction), STEM, and medicine that require precise 3D spatial understanding. The impact of spatial visualization skill is both educational and practical.
\par
AI with spatial reasoning skills can greatly benefit educational and practical tasks.  In education, despite having access to various computer-assisted learning resources for geometry, spatial transformations, and mathematical concepts, many STEM students struggle with visualization tasks \cite{garmendia2007first},  \cite{andrews2017hypothetical}. When conversational AI is added to an immersive visual learning approach, learners can receive personalized, real-time assistance, allowing them to ask questions, correct misunderstandings, and practice spatial reasoning interactively \cite{wei2025multiple}.
By leveraging tools such as AR, virtual labs, and AI-driven assistance, students can explore STEM topics at their own pace and develop critical problem-solving skills in an interactive, engaging way \cite{monjoree_ar-classroom_2024}.
In practice, in construction, manufacturing, disaster response, and emergency field operations, AI's understanding of spatial relationships can facilitate assembly, avoid misplacement of components, help navigation, and facilitate prompt response.

\par

Generative AI, encompassing technologies that can create new content such as text, images, 3D models, audio, and video, has rapidly evolved in recent years, driven by advancements in deep learning and neural network architecture. Large language models (LLMs) such as GPT-4 and its successors have demonstrated remarkable proficiency in generating coherent and contextually relevant text. GPT-4o and GPT-5 with vision have capabilities to analyze user-input images and show unprecedented ability in understanding visual inputs, spatial relationships of humans and objects \cite{achiam2023gpt},\cite{han2026decoupling}.

Generative AI's visual capabilities are increasingly relevant to augmented reality (AR), virtual design, construction guidance, disaster response, and other applications that require real-time understanding of visual and spatial input. However, current vision-language models still have limitations in spatial reasoning, especially when interpreting three-dimensional transformations such as rotation, translation, scale, orientation, and perspective changes \cite{han2020spare3d}. This creates a critical knowledge gap: AI systems can describe images and objects, but they are not yet consistently reliable as spatial guides in complex real-world environments where decisions may affect safety, infrastructure, and human life \cite{anthi2025role}, \cite{lee2025agentic}. Therefore, improving AI’s spatial intelligence is essential for building trustworthy human-AI interaction systems that can support users in dangerous, high-pressure, and visually complex settings \cite{speiser2026agentic}. 

\par

Spatial reasoning, a key area in psychometrics, has several datasets for evaluating these skills. Some widely used tests for rotation visualization are the Purdue Spatial Visualization Test (PSVT)/The Purdue Visualization of Rotations Test \cite{bodner1997purdue}, the Revised Purdue Spatial Visualization Test: Visualization of Rotations (Revised PSVT:R) \cite{yoon2011psychometric}, and the Mental Rotation Tests \cite{vandenberg1978mental}. The Revised PSVT:R is a popular method to measure an individual's ability to visualize and mentally rotate 3D objects in space. 
\par
We can use both a verbal and visual description to explain spatial rotations and convey how objects change orientations in 3D space \cite{lacey2013multisensory}. It is possible to improve AI's spatial rotation understanding by combining the visual and verbal abilities of conversational AI \cite{monjoree_ar-classroom_2024}, \cite{11050695}. Previous studies have shown that Vision Language Models (VLMs) struggle with spatial reasoning tasks despite using methods like fine-tuning and Chain-of-Thought (CoT) \cite{xu-etal-2025-defining}. The aim of this study is to enhance AI’s capability of understanding 3D spatial rotations by training currently available VLMs through an Agentic AI approach and performing a series of tests based on Revised PSVT:R. The research questions that guide this study are as follows:

\begin{enumerate}
    \item What is an efficient process for improving generative AI models' performance in Revised PSVT:R using an Agentic AI system?
    \item How does the performance of Agentic AI compare with the previously trained models and human test-takers?
    
\end{enumerate}

\section{Literature Review}

\subsection{AI’s Spatial Abilities}

Spatial intelligence aids us in comprehending visual-spatial tasks and spatial relations. Thus, we gain a better orientation of objects in space \cite{hawes2022effects}. Recently, there have been studies analyzing the spatial and mathematical capabilities of AI. The SPARE3D dataset reveals that while convolutional neural networks excel in visual learning, their performance on spatial reasoning remains below human average. This article highlights the need for advancements in artificial neural network design to improve 3D spatial reasoning for intelligent robots \cite{han2020spare3d}. Research indicates that while GPT-4, Gemini 1.5 Pro, and Llama 3.2, as the main current generative AI models, lack the understanding of a spatial rotation process, it has the potential to understand the rotation process when language (mathematical equation describing the rotation) is utilized to process spatial information \cite{11050695}. Basic spatial abilities tests demonstrated that VLMs mirror human hierarchies (strongest in 2D orientation, weakest in 3D rotation) \cite{xu-etal-2025-defining}. Moreover, interventions like Chain-of-Thought (CoT) and few-shot training showed limitations, while Tree-of-Thoughts (ToT) demonstrated the most effective enhancement \cite{xu-etal-2025-defining}.

A recent study has examined the perceived complexity of 3D shapes from both human and ChatGPT perspectives, revealing that ChatGPT could be used to automatically generate 3D content with varying complexities, which can be enhanced for spatial visualization skills in VR applications \cite{mcshane2023perceived}.  
After an elaborate study, researchers observed that state-of-the-art LLMs resemble sophisticated pattern matching more than true logical reasoning\cite{mirzadeh2024gsm}. LLM performance varies highly on different versions of the same question and a minor increase in difficulty. Another study found that ChatGPT-4 and ChatGPT-4o struggle more with geometry and complex math problems compared to algebra, suggesting that while AI can support algebra instruction, teacher intervention is needed in areas where AI is less effective, such as spatial intelligence and higher-difficulty tasks \cite{wei2024evaluating}. Overall mathematical performance of AI falls significantly short of a graduate student’s level \cite{frieder2024mathematical}.
\par
Despite the drawbacks, in recent years, AI has demonstrated the potential to reshape education by enhancing interactivity, personalization, and accessibility. AI and AR-based multi-sensory assistance helps undergraduate students better understand spatial transformations and their mathematical representations \cite{monjoree_ar-classroom_2024}.
\subsection{VLM Applications}

Large-scale, multi-modal vision-language AI models can process image and text inputs and produce text outputs. Although they are still less capable than humans in practical usage, GPT-5 Vision, Gemini 3.1 Pro, and Llama 4 can identify objects, analyze images, provide detailed descriptions, and offer explanations of scenes \cite{hessenthaler2026evaluating}. AI with vision and spatial understanding has a wide range of applications in STEM as well as AEC, such as construction safety management, risk assessment, real-time insights, incident reporting, and compliance evaluation \cite{saka2024gpt}, \cite{chen2025tailored}, \cite{zhang2026cognitively}. Experiments have proven the GPT vision model to be highly effective in detecting and explaining global and fine-grained semantic patterns in anomaly or risk detection \cite{yang2023dawn}, \cite{driessen2024putting}.

\subsection{AI in AEC Training and Practice}

Experimental results show that GPT-4o and Gemini 1.5 Pro have the ability to identify both general and context-specific hazards even though longer processing times impact real-time feasibility \cite{adil2025using}. Fine-tuned VLMs for construction hazard report generation can automatically produce structured and detailed hazard reports and provides an innovative solution for construction safety management, enhancing efficiency, accuracy, and automation in construction hazard identification \cite{chen2025tailored}. Methodical prompt modularization and hybrid optimization of VLMs showed professional-level performance in fire-safety tasks, providing a scalable and practical computational pipeline for VLM-assisted urban building safety supervision \cite{zhang2026cognitively}.

\subsection{Agentic AI}

Agentic AI generates responses for autonomous systems that can independently plan to achieve goals with minimal human intervention. These models can observe, adapt, coordinate with other agents, and even refine their own outputs over time \cite{hosseini2025role}, \cite{khalid2026agentic}. Agentic AI promotes workflow optimization and goal-oriented processes by combining language understanding, reasoning, planning, and decision-making. Recent studies show the use of Agentic AI in the construction process. Autonomous AI agents, powered by LLMs, can utilize multi-modal sensing tools to autonomously identify, classify, and localize structural damage \cite{11596258}. GPT-based AI Agent can function as a post-correction mechanism and provide a scalable, automated alternative to spreadsheet-based workflows, enabling improved bidding, project planning, and business intelligence \cite{kalasapudi2025agentic}. In this paper, Agentic AI is used with a goal of finding the correct answers for the Revised PSVT:R test questions, through reasoning, using the CoT and tools (e.g., for text and image reading, image slicing, table creation, among others).

\subsection{Dual-coding theory}

The dual-coding theory defines the existence of two interconnected subsystems in the human cognitive system: an imagery system and a verbal system  \cite{clark1991dual}. The dual capacity shows that the language system functions in both symbolic interpretation and communication. Human cognition integrates both language and nonverbal objects and events, which enables us to process linguistic input (i.e., writing and speech) with symbolic representations of nonverbal aspects of the world \cite{paivio1990mental}. Hence, spatial information can be represented through integrating these two systems, where language serves both as direct communication and as a framework for understanding nonverbal phenomena. However, studies also demonstrated that linguistic representations are neither utilized nor required for certain inductive or deductive logical reasoning \cite{kean2026evidence}. Other studies suggest that natural language enables us to express complex, structured meanings \cite{guttenplan1975thought, munroe2022you}. The spatial visualization that this paper concerns presents a significantly different reasoning task that requires interpreting 3D geometry and its spatial transformations - whether AI as a test-taker may benefit from a linguistic representation in the reasoning process is being investigated.

\section{Methodology}

Dual-coding describes how two cognitive subsystems, imagery and verbal, work together to process language and non-verbal content. When people explain rotation, they use references like direction, angle, and axis visually and verbally. Integrating visual and verbal systems into the Agentic AI training process has the potential to enhance spatial understanding, where language aids in communication and vision aids in interpreting nonverbal cues.  \par
LLMs rely on descriptive datasets and visual cues, which can enhance spatial visualization in vision models. Building on prior work of evaluating AI models \cite{11050695}, \cite{monjoree2024aisspatialintelligenceevaluating}, we focused on enhancing VLMs' spatial ability by utilizing an agentic AI system. In this work, Agentic AI utilizes a conventional VLM (using GPT-5.6 high reasoning mode) by enabling systems to plan, make decisions, and interact with external environments or tools. The Agentic AI iteratively takes actions toward achieving the specified goal of perceiving 3D object rotation and addresses complex multi-step tasks of Revised PSVT:R (Fig. \ref{fig1}) with reduced human intervention. In this paper, the agentic AI system aims to identify the correct answer by employing chain-of-thought (CoT) reasoning and specialized tools for image interpretation, image slicing, table creation, and related tasks.

CoT prompting is a key reasoning technique that encourages VLMs to generate intermediate reasoning steps before producing a final answer. Research has demonstrated that CoT prompting can substantially improve performance on arithmetic, commonsense, and symbolic reasoning tasks \cite{wei2022chain}.
\par 
We used both structured CoT and structured CoT with self-optimization to improve the performance of the AI model for understanding rotation. Structured CoT extends conventional CoT prompting by organizing intermediate reasoning through imposing task-relevant structure rather than allowing the model to produce an unconstrained sequence of reasoning steps \cite{li2025structured}. In the self-optimization process, the model evaluates, selects, refines, and dynamically controls and compresses reasoning according to task characteristics \cite{huang2025reasoning}.

\par
In this paper, we used OpenAI GPT-5 Codex CLI \cite{openai_codex_cli}) for all our experiments. In these experiments, we are relying on the AI model's vision capability to identify visual cues and text on images representing the rotation process. Previous study demonstrated that superimposing the axes system in the Revised PSVT:R dataset was not statistically significant when tested on base VLMs (GPT-4, Gemini 1.5 Pro, and Llama 3.2) \cite{11050695}. For this reason, we are using CoT-structured CoT prompting along with the axes system to improve AI's spatial visualization performance (Fig. \ref{fig2}).

\par

\section{Experiments }

\begin{figure}[htbp]
\centerline{\includegraphics[width=1\linewidth]{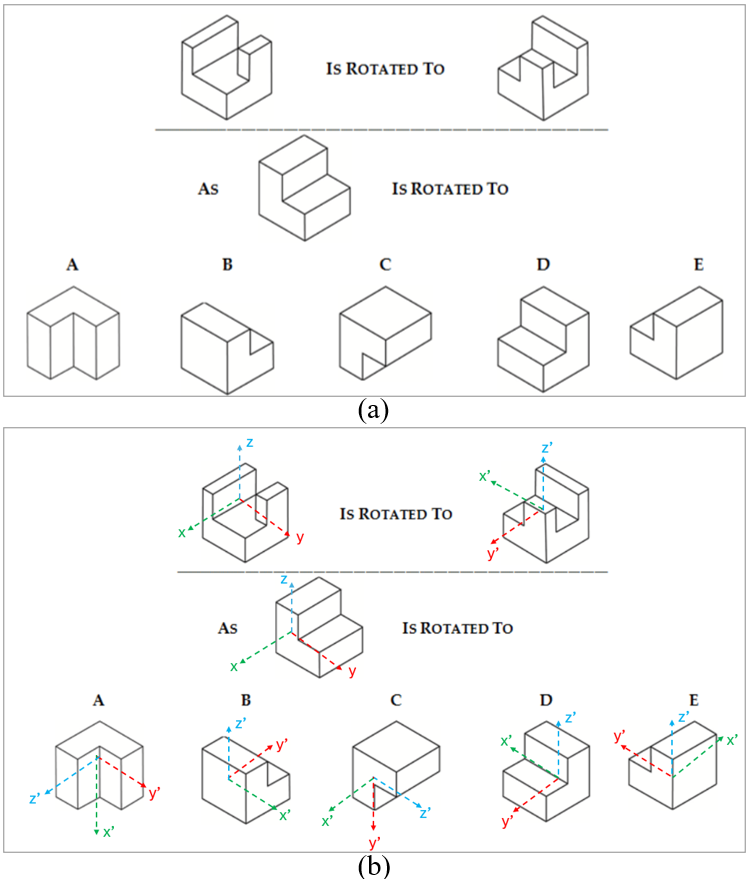}}
\caption{(a) An example question from Revised PSVT:R, (b) the example question from Revised PSVT:R w/ Axes. In both questions, D is the correct answer.}
\label{fig1}
\end{figure}

\begin{figure*}[t] 
    \centering
    \includegraphics[width=.9\textwidth]{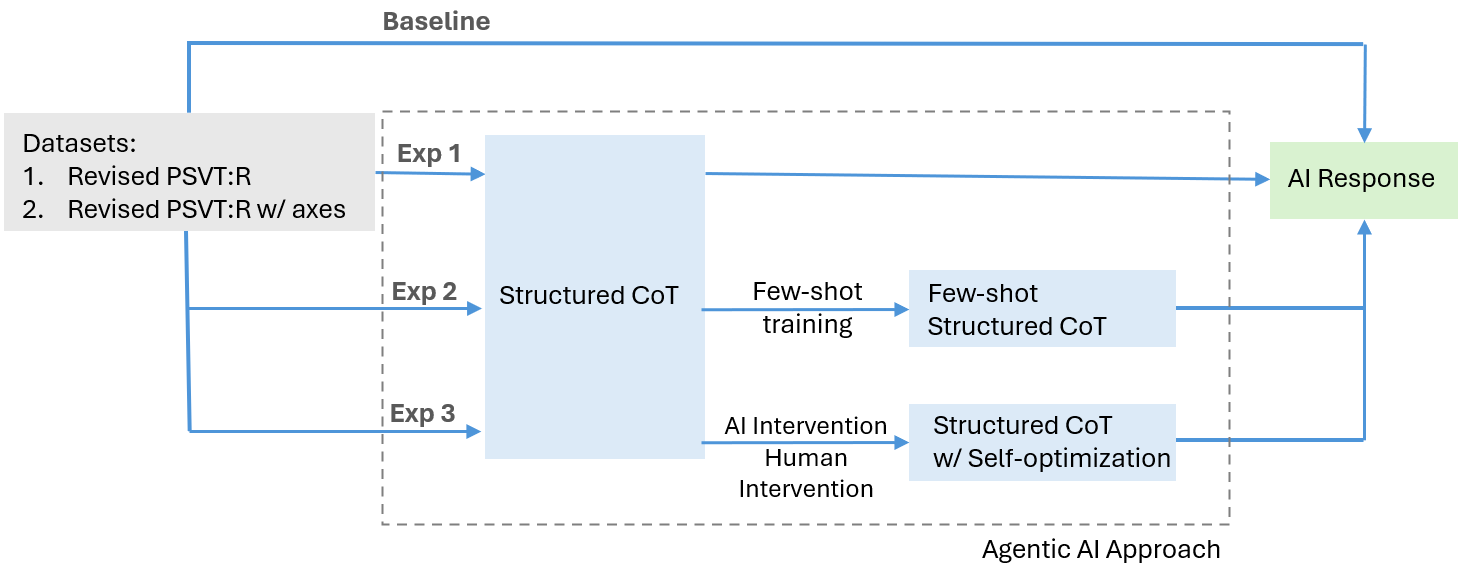}
    \caption{Framework of Baseline Test and Three CoT Experiments}
\label{fig2}
\end{figure*}

The Revised PSVT:R consists of 30 rotation problems \cite{yoon2011psychometric}. It is an extensively used test for assessing spatial visualization skills. The test-takers in this test are shown a figure of 3D objects in both starting and finishing orientations after an unknown rotation along one or more axes. Their task is to identify the correct figure of a second 3D object that went through the same rotation process as the first 3D object from five possible options. In our study, both test sets (Revised PSVT:R and Revised PSVT:R with a coordinate system - Revised PSVT:R w/ Axes (Fig. \ref{fig1})) will go through a baseline test and three experiments: (1) structured CoT prompting, (2) few-shot structured CoT prompting, and (3) structured CoT with self-optimized prompting.

\textbf { Prior Work: }In a prior study, VLMs showed inadequate performance when tested using the Revised PSVT:R dataset \cite{11050695}. GPT-4, Gemini 1.5 Pro, and Llama 3.2 showed 16.7\%, 23.3\%, 16.7\% accuracy in Revised PSVT:R test. In this paper, we used a newer version of VLM and utilized the structured CoT method to improve those results.

\textbf { Baseline Test Using Revised PSVT:R and Revised PSVT:R w/Axes:} At first, we provided GPT-5.6v (high reasoning) with all 30 problems from the Revised Purdue Spatial Visualization Test: Visualization of Rotations (Revised PSVT:R). It is considered a standard test to study its ability to analyze input images and spatial rotation capabilities (Fig. \ref{fig1}a). After receiving the uploaded input (a screenshot of the test questions), the generative AI model responded by answering the questions by choosing from the (A, B, C, D, and E) options. To keep the test results confidential for the copyright of Revised PSVT:R and for future tests with participants, the 30 questions and results are not demonstrated in this paper. We show an example question from the DIRECTIONS section of the test in Fig. \ref{fig1}(a).
\par
We also used the test with a coordinate system with x-, y-, and z-axes on the Revised PSVT:R question images, named as Revised PSVT:R w/ Axes (Fig. \ref{fig1}b). The purpose of this dataset is to test whether more visual and textual cues can improve communication with AI and its performance.
\par

\textbf{Exp 1: Structured CoT Prompting}
\par
\begin{figure}[htbp]
\centerline{\includegraphics[width=\linewidth]{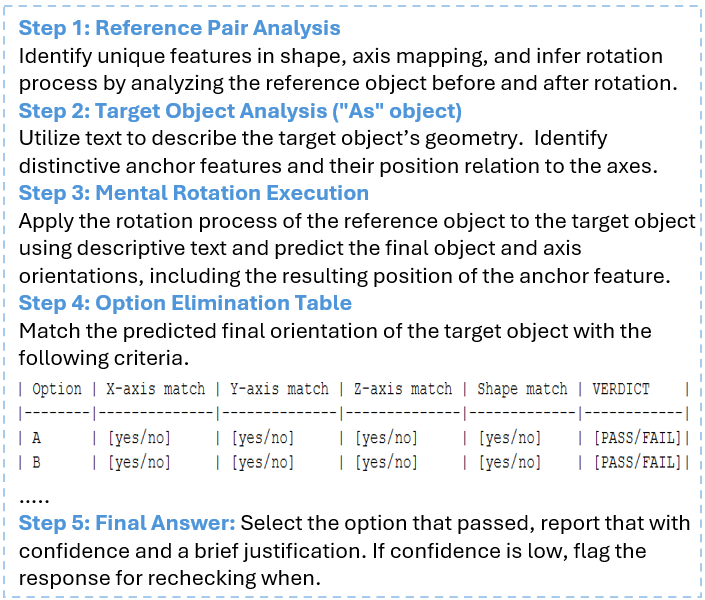}}
\caption{Structured CoT}
\label{fig3}
\end{figure}

In this section, we developed an elaborate, structured CoT for the Agentic AI to process. Here, we broke down the rotation perception task into smaller and simpler steps to make the test question easier to comprehend and imitate the thought process a human test-taker would follow (Fig. \ref{fig3}). In this process, the Agentic AI is instructed to infer the axis system according to the CoT instruction where the axes are not given (30 questions of Revised PSVT:R). In the case of the second dataset PSVT:R w/ Axes, Agentic AI will utilize the axes. Step 1 focuses on understanding the reference object, then finding and describing its unique features, such as ``notch on right face" and ``cavity on left". After that, Agentic AI finds the same features in the rotated reference object and then describes the rotation process. In step 2, it finds the unique features of the target object. Then, in step 3, Agentic AI applies the rotation of the reference object to the target object and predicts where the unique feature would be verbally. In step 4, it matches its predicted final orientation of the target object using criteria in the table and provides a verdict for each option (Fig. \ref{fig3}). In the last step, it provides an answer with the confidence level. For detailed structured CoT, see (Fig. \ref{fig:A1})in the Appendix. For an AI-generated sample answer, see (Fig. \ref{fig:A2}) in the Appendix.
\par

\textbf{Exp 2: Few-shot Structured CoT Prompting}
\par
LLMs show significant capabilities for zero-shot. However, they still struggle with complex tasks when using the zero-shot setting \cite{kim2023cot}. For this reason, we used few-shot prompting with structured CoT (Fig. \ref{fig3}) as a technique to enable in-context learning. Here, we provided 5 questions from Revised PSVT:R with detailed solutions for training purposes in the prompt to direct the model to perform better. The examples serve as conditioning for subsequent examples where we would like the model to generate a response. We used the remaining 25 questions for testing purpose in this experiment. The CoT prompt is similar to (Fig. \ref{fig:A1})in the Appendix. For an example of a detailed answer from AI, see (Fig. \ref{fig:A2}) in the Appendix.
\par

\textbf{Exp 3: Structured CoT with Self-optimized Prompt}

\par
\begin{figure}[htbp]
\centerline{\includegraphics[width=.8\linewidth]{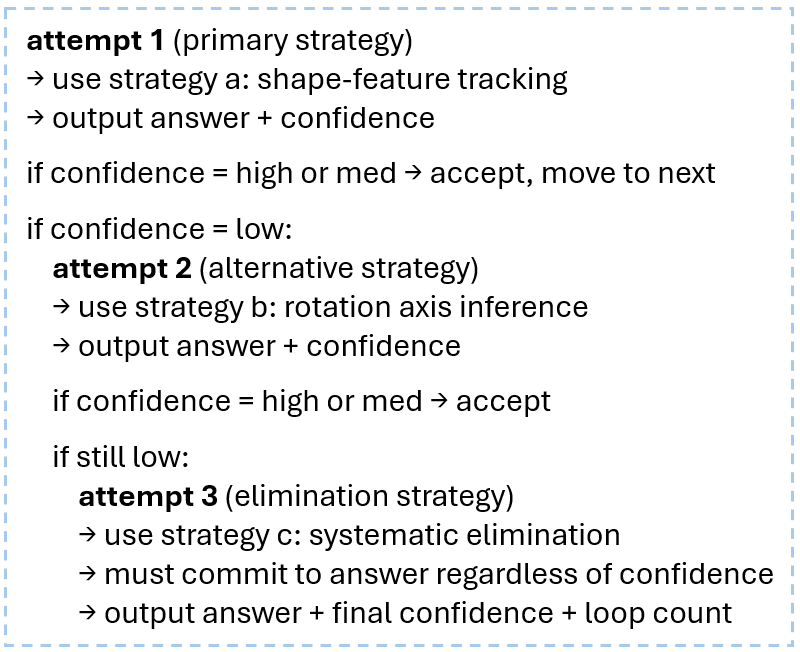}}
\caption{Structured CoT with Self-optimized Prompt }
\label{fig4}
\end{figure}

In this experiment, we provided Agentic AI with our structured CoT and instructed Agentic AI to self-optimize with a goal of achieving an improved understanding of the rotation process. It proposed a combination of three consecutive strategies to perceive the rotation process (Fig. \ref{fig4}). These three strategies were evaluated to check logical reasoning to keep the human in the loop. The Agentic AI follows three strategies one after another if the confidence level is low. After going through each strategy, Agentic AI provides an answer and rates its confidence level. If the confidence level is high or medium, it finalizes the answer. If the confidence level is low, the next strategy follows. AI answered the 30 questions from the Revised PSVT:R.

\begin{enumerate}[label=(\alph*)]
    \item Shape-Feature Tracking
    \item Rotation Axis Inference
    \item Systematic Elimination
\end{enumerate}

\begin{figure}[t]
    \centering

    \begin{subfigure}{\columnwidth}
        \centering
        \includegraphics[width=\columnwidth]{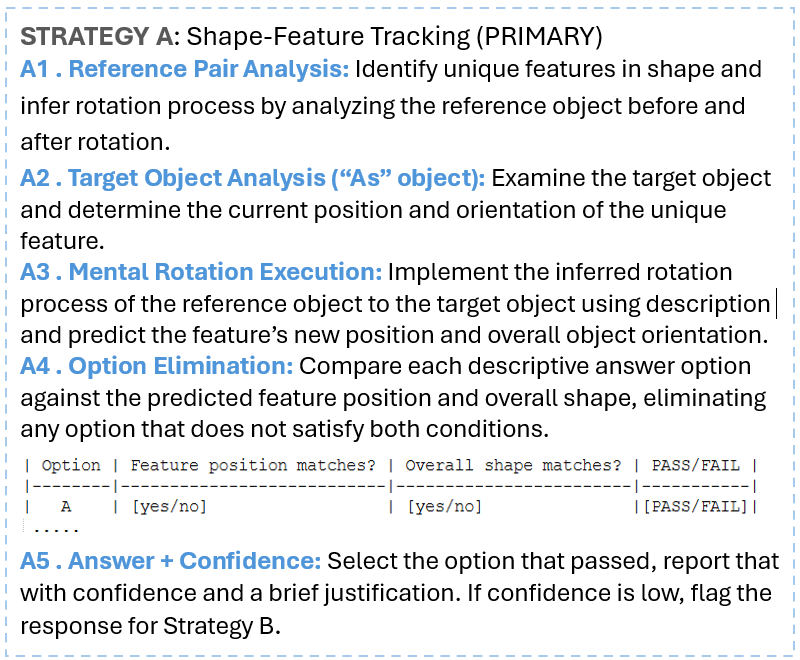}
        \caption{}
        \label{fig:image_a}
    \end{subfigure}

    \vspace{3pt}

    \begin{subfigure}{\columnwidth}
        \centering
        \includegraphics[width=\columnwidth]{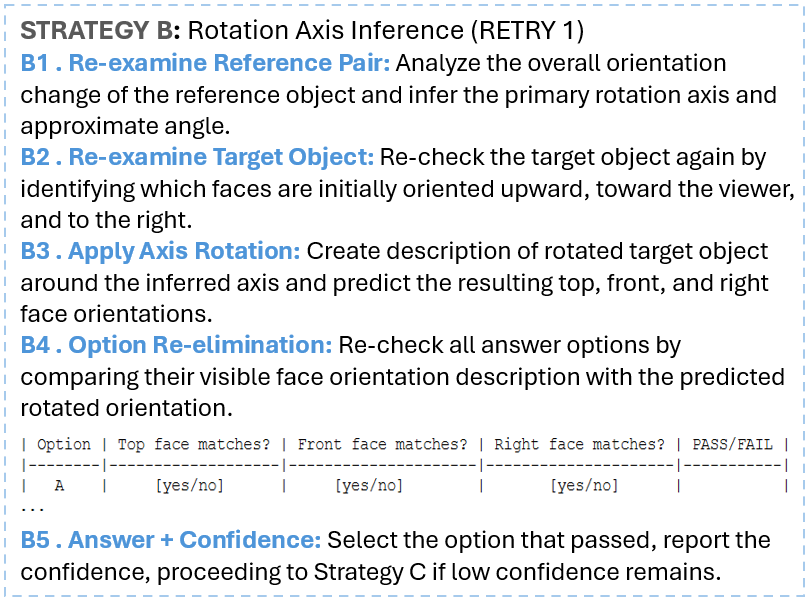}
        \caption{}
        \label{fig:image_b}
    \end{subfigure}

    \vspace{3pt}

    \begin{subfigure}{\columnwidth}
        \centering
        \includegraphics[width=\columnwidth]{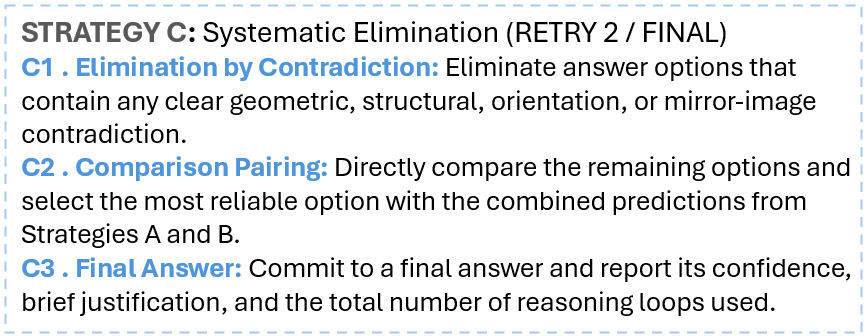}
        \caption{}
        \label{fig:image_c}
    \end{subfigure}

    \caption{Three Strategies of Structured CoT with Self-optimized Prompting}
    \label{fig5}
\end{figure}
\par
\textbf{\textit{Strategy A: Shape-feature Tracking}}
\par
The Strategy A is similar to our structured CoT. The axis system is utilized in the second strategy. Step A1 (Fig. \ref{fig5}a) of the Strategy A focuses on understanding the reference object and finding and describing its unique features, such as "protruding block on top-left" and "open cavity faces left". Following that, Agentic AI identifies the same features in the rotated reference object and then describes the rotation process verbally. Step A2 focuses on the unique features of the target object. In step A3, Agentic AI applies the rotation of the reference object to the target object and predicts verbally where the unique feature would be. In step A4, Agentic AI uses criteria in the table (Fig. \ref{fig5}a) and provides a verdict for each option. Finally, in step A5, it provides an answer with a confidence level. For the detailed instructions for Strategy A, see (Fig. \ref{fig:A3}) in the Appendix.
\par
\textbf{\textit{Strategy B: Rotation Axis Inference}}
\par
In Strategy B, the rotation process is defined by the axis system (Fig. \ref{fig5}b). Agentic AI assumes an axis system even if it is not provided in the image. In step B1, Agentic AI re-examines the reference pair by analyzing the overall orientation change of the reference object and infers the primary rotation axis and approximate angle. In step B2, Agentic AI re-checks the target object again by identifying which faces are initially oriented upward, toward the viewer, and to the right. Step B3 focuses on creating a description of the rotated target object around the inferred axis and predicting the resulting top, front, and right face orientations. Step B4 re-checks all answer options by comparing their visible face orientation description with the predicted rotated orientation. In the final step B5, one option is selected with a confidence level, proceeding to Strategy C if low confidence remains. For the detailed instruction of strategy B, see (Fig. \ref{fig:A4}) in the Appendix.
\par
\textbf{\textit{Strategy C: Systematic Elimination}}
\par
Strategy C (Fig. \ref{fig5}c) focuses on a systematic elimination process, where step C1 eliminates answers found in strategies A and B for the reason of contradiction. Here, Agentic AI eliminates answer options that contain any clear geometric, structural, orientation, or mirror-image contradiction. Then, in step C2, it directly compares the remaining options and selects the most reliable option with the combined predictions from strategies A and B. Finally, in C3, the most probable answer is given with its confidence and a brief justification. If the confidence level is still low, it goes through another loop of strategies A, B, and C.
At the end, the total number of reasoning loops used. For the detailed instruction of Strategy C, see (Fig. \ref{fig:A5}) in the Appendix.
\par
Exp 1, 2, and 3 (Strategy A, B, and C) have been applied to Revised PSVT:R and Revised PSVT: R w/ Axes. For an example of a detailed answer from AI, see (Fig. \ref{fig:A6}) in the Appendix.

\textbf{}  

\par

\section{Results and Discussion}

VLMs are effective multimodal tools that interpret images alongside text descriptions. While LLMs are more accurate for text-based tasks, the vision models' accuracy for purely image-based tasks, like object detection or more complex image analysis, may not match that of specialized vision models \cite{team2024gemini} and still rely heavily on textual inputs to guide their performance. We can observe from Table 1 that AI performed better in Revised PSVT:R w/ Axes than Revised PSVT:R for the Baseline Test and Exp 1, Exp 2, and Exp 3. Also, we can notice that using the structured CoT method for Exp 1, Exp 2, and Exp 3 has increased the performance of the AI from the Baseline for both datasets. 

\begin{table*}[!t]
\centering
\caption{Performance comparison based on PSVT:R with percentage of correct answers.}
\label{tab1}

\setlength{\tabcolsep}{4pt}

\begin{tabularx}{\textwidth}{
    @{}
    P{2.3cm}
    L
    L
    P{1.4cm}
    P{1.4cm}
    P{1.4cm}
    P{1.4cm}
    @{}
}
\toprule

& \multicolumn{2}{l}{\textbf{Performance of Human Test-Takers}}
& \multicolumn{4}{l}{\textbf{Performance of AI}} \\

\cmidrule(lr){2-3}
\cmidrule(lr){4-7}

\textbf{Test Condition}
& \textbf{Study 1 \cite{sorby2009educational}}
& \textbf{Study 2 \cite{yoon2011psychometric}}
& \textbf{Baseline}
& \textbf{Exp 1}
& \textbf{Exp 2}
& \textbf{Exp 3} \\

\midrule

\textbf{Original or Revised PSVT:R}
& 18\% participants scored 60\% or less; 82\% scored above 60\%
& Participants' mean accuracy was 63.60\%, participants' accuracy ranged from 10\% to 100\%
& 13.33\%
& 40\%
& 48\%
& 43.33\% \\

\addlinespace

\textbf{Revised PSVT:R w/ Axes}
& N/A
& N/A
& 26.67\%
& 100\%
& 96\%
& 96.70\% \\

\bottomrule
\end{tabularx}
\begin{tablenotes}
\footnotesize
\item \textit{Note:} AI was instructed not to search for answers on the Internet in the  experiments for the Revised PSVT:R tests as the goal was to evaluate AI's spatial reasoning.
\end{tablenotes}
\end{table*}

\textbf{Performance of Agentic AI vs Human Test-takers:} In Study 1 from 1993 (Table \ref{tab1}), a PSVT:R was administered to 535 first-year engineering students at Michigan Tech. In that test, around 18\% of the students (n = 96) scored 60\% or less on the PSVT:R, while 82\% of the students scored above 60\% accuracy \cite{sorby2009educational}. In Study 2 published in 2011 (Table \ref{tab1}), the performance of human test-takers (undergraduate students) in the Revised PSVT:R showed that the mean score of 1,022 participants was 19.08 out of 30 (63.6\% mean accuracy), with a standard deviation of 6.16 \cite{yoon2011psychometric}. Scores ranged from 3 to 30 (10\% to 100\% accuracy), and 13 participants (1.27\%) achieved a perfect score. We can see from Table \ref{tab1} that the base model (GPT 5.6 with high reasoning ability) underperformed humans in this spatial visualization - PSVT:R (Baseline). Even though the Agentic AI performed much better than the basic AI model, it still underperformed humans in this spatial visualization - Original/Revised PSVT:R (Exp 1, Exp 2, and Exp 3). However, the Agentic AI in Exp 1, Exp 2, and Exp 3 performed better than underperforming participants for Revised PSVT:R (lowest accuracy being 10\% in Study 2).

\textbf{Performance of Agentic AI Based on Three CoT Experiments:}
\par
\textbf{\textit{Comparison of Baseline with Experiment 1, 2, and 3:}} A Chi-square test of homogeneity at p$<$0.05
indicates that the performances of Baseline and the structured CoT experiments (Exp 1, Exp 2, Exp 3) have significant group differences for Revised PSVT:R and Revised PSVT:R with axes. Baseline, Exp 1, Exp 2, and Exp 3 had 30, 30, 25, and 30 samples, respectively.
\begin{enumerate}
    \item [(a)] $\chi^2$  (3, N = 115) = 9.194, \textit{p} $=$ .0268 for Revised PSVT:R
    \item [(b)] $\chi^2$  (3, N = 115) = 53.2, \textit{p} $<$ .001 for Revised PSVT:R w/ Axes
\end{enumerate}
\par These results indicate that all structured CoT methods have improved AI's performance significantly over the base GPT-5.6 model.
\par
\textbf{\textit{Comparison of Experiments 1, 2, and 3:}}
There are some performance differences among the three approaches that we took in this paper: (Exp 1) structured CoT prompting, (Exp 2) few-shot and structured CoT prompting, and (Exp 3) structured CoT with self-optimized prompting (Table \ref{tab1}). However, the Chi-square test of homogeneity indicates that there is no statistically significant group difference among the performances of the three approaches (p$>$.05) for Revised PSVT:R and Revised PSVT:R w/ Axes. Using different structured CoT methods (in Exp 1, Exp 2, and Exp 3) did not make any significant group differences. Exp 2 and Exp 3 are modified versions of Exp 1 to learn if the performance of structured CoT could be enhanced. However, incorporating the few-shot technique for Exp 2 and self-optimizing prompts for Exp 3 did not make a significant performance difference, respectively, compared with Exp 1. \par

\textbf{\textit{Comparison of Revised PSVT:R with Revised PSVT:R w/ Axes:}} The results of Table \ref{tab1} also indicate that the spatial intelligence of the current generative AI models does not perform well without additional context information (for example, axes). For CoT methods, a Chi-square test of homogeneity at p$<$0.05
indicated a statistically significant group difference between Revised PSVT:R and Revised PSVT:R w/ Axes for Exp 1, Exp 2, and Exp 3. Exp 1, Exp 2, and Exp 3 had 30, 25, and 30 samples, respectively.
\par $\chi^2$  (1, N = 85) = 91.5, \textit{p} $<$ .0001. \par However, for the Baseline tests, there is no significant group difference between Revised PSVT:R and Revised PSVT:R w/ Axes.\par $\chi^2$  (1, N = 30) = .938, \textit{p} $=$ .3329 \par 

Additional context (axis lines and axis names) in the image significantly improves Revised PSVT:R question analysis using language, and consequently, the performance of structured CoT approaches improves. These results also demonstrate conformity with the dual-coding theory, which posits that human cognition integrates both language and nonverbal objects to process linguistic input with symbolic representations of nonverbal aspects of the world. Here, we provided a textual and graphical reference (axis lines and axis names) in the image and divided a complex visual analysis task in simple, smaller steps with linguistic instructions for Agentic AI to process. As a result, the combination of structured CoT and contextual information in the image showed very promising results for Exp 1, Exp 2, and Exp 3 in Revised PSVT:R w/ Axes (Table 1). In the future, this process has the potential to help train AI in deep learning to improve its spatial intelligence. \par
\textit{Note:} PSVT:R questions have been used in several studies throughout the years. Those may or may not have been used to train the base GPT-5.6 model. To investigate how AI performs if it takes hints (i.e. direct, correct answers for the PSVT:R test questions) from available resources, in another experiment, we specifically asked AI to utilize internet sources (such as Hugging Face datasets \cite{xu2025basicspatialabilitytest}, which actually have the complete PSVT:R questions and answers) to find answers for the Revised PSVT:R with the Exp 1 structured CoT. The accuracy rate of that experiment was 100\%, as expected. It mentioned a Hugging Face dataset \cite{xu2025basicspatialabilitytest} as the source of the correct answers that was created for the purpose of testing AI's spatial ability \cite{xu-etal-2025-defining}. However, that means AI acquired the answers instead of conducting reasoning to get the answers. In comparison, in all experiments presented earlier in this paper, AI was asked specifically not to look for existing answers. Based on the output of both the various accuracies of the answers and the verbiage reasoning block (e.g., Fig. \ref{fig:A2}), we verify the AI's reasoning process, which is the subject of this research.

\par
AI with spatial intelligence can be beneficial in the future in education and practice. For example, after generative AI models learn to understand spatial transformations (such as rotations) and their mathematical representations, the models could guide students step by step, by asking students questions on the topics and offering real-time feedback in a  class on rotation axes, angles, rotation sequences, transformation matrices, etc., enabling more effective learning by the students regarding spatial transformations and their mathematical representations. 

\section{Conclusions and Future Work}

Spatial skills are critically important in STEM and art (STEAM) fields like architecture, computer graphics, civil engineering, medicine, manufacturing, etc. Enhanced AI's spatial intelligence is expected to improve immersive AR/VR-assisted manufacturing. This paper investigates an agentic AI approach to enhance VLMs' spatial capabilities, particularly their proficiency in interpreting 3D rotations. We improvised the Revised PSVT:R with additional graphical and textual features to examine the performance changes in different types of CoT methods. We learned that the structured CoT method improves the spatial performance of the base GPT-5.6 model for both datasets significantly, even though there are no significant performance differences between the three CoT approaches. While in the absence of contextual information, agentic AI relying solely on CoT reasoning continues to exhibit limitations in spatial reasoning capabilities, the reasoning of structured CoT and contextual information together for analyzing 3D rotation in the experiments showed very promising results, which can be used to inform future AI training. 
\par
\par
As we utilized the GPT model with high reasoning ability, it took significant time to generate answers. Future research and AI development may resolve this issue. Future work will include testing with new and larger datasets (i.e., questions and answers created based on the Revised PSVT:R purpose), the latest AI models, fine-tuning AI models with the CoT method, etc. An initial test of the latest GPT-6 Astra Max model revealed that the baseline model didn't provide a reasoning accurately for a sample Revised PSVT:R question, indicating a continued need for testing the latest AI models and enhancing their spatial reasoning potentially using the proposed method from this paper. Additionally, we can enhance the spatial intelligence of AI by training with rotations, translations, scales, perspectives, etc images. Using different styles of image or video input of arbitrary 3D physical/digital objects, with or without contextual information of axes, text, math formulas, etc., can be beneficial.

\section*{Acknowledgment}

This work is supported by the National Science Foundation (NSF) Grant No. 2119549 and the NSF RETTL project team at Texas A\&M University.

\bibliographystyle{ieeetr}
\bibliography{refs}

\clearpage
\onecolumn
\raggedbottom
\appendix
\renewcommand{\thefigure}{A\arabic{figure}}
\setcounter{figure}{0}
Find the detailed CoT prompting used in the experiments below.

\begin{figure}[H]
    \centering
    \includegraphics[width=.9\textwidth]{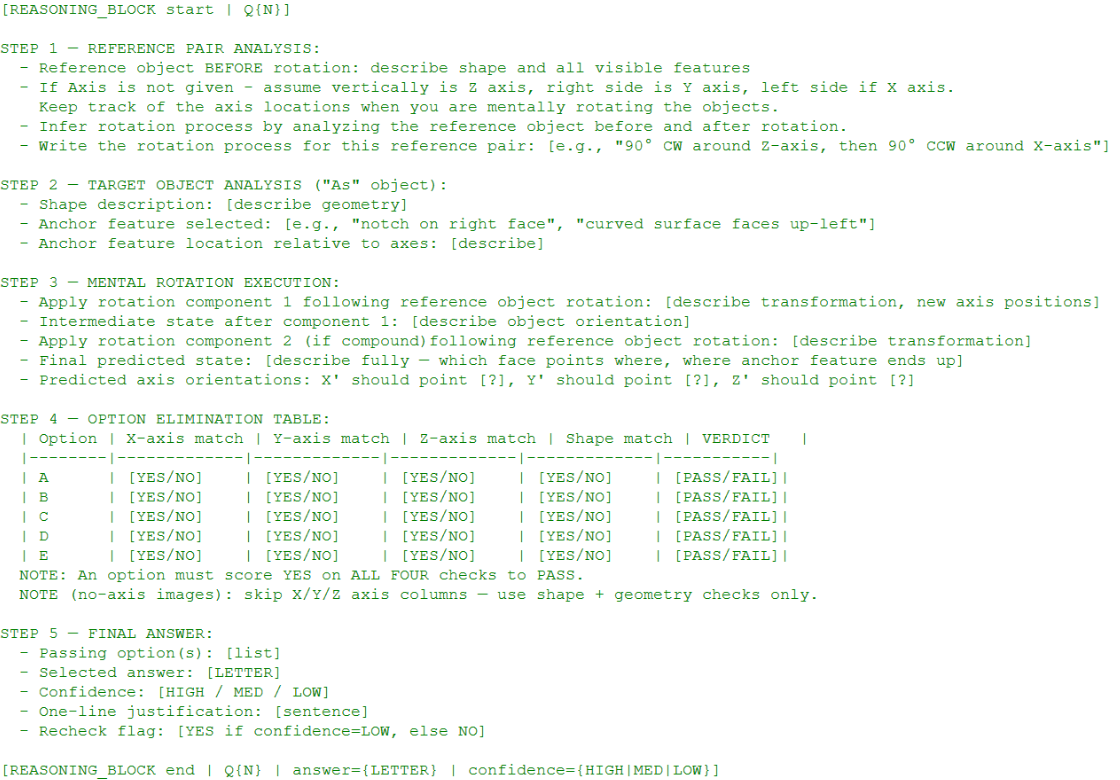}
    \caption{Prompt for Structured CoT used in Experiment 1 \& Experiment 2}
    \label{fig:A1}
\end{figure}

\begin{figure}[H]
    \centering
    \includegraphics[width=1\textwidth]{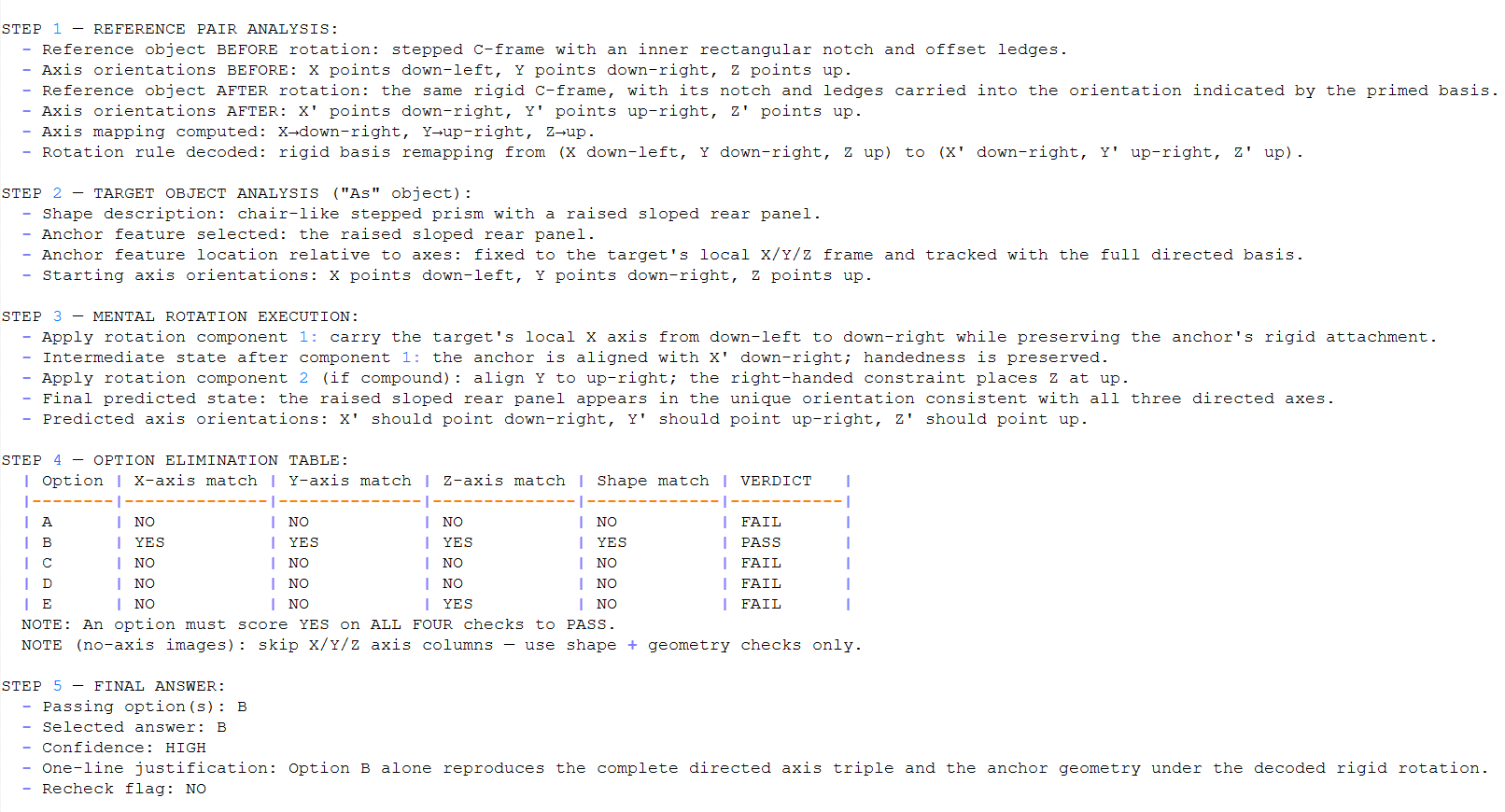}
    \caption{Sample answer of the Structured CoT method used in Experiment 1 \& Experiment 2. This is the answer to the question of Revised PSVT:R (Correct answer B)}
    \label{fig:A2}
\end{figure}

\begin{figure}[H]
    \centering
    \includegraphics[width=1\textwidth]{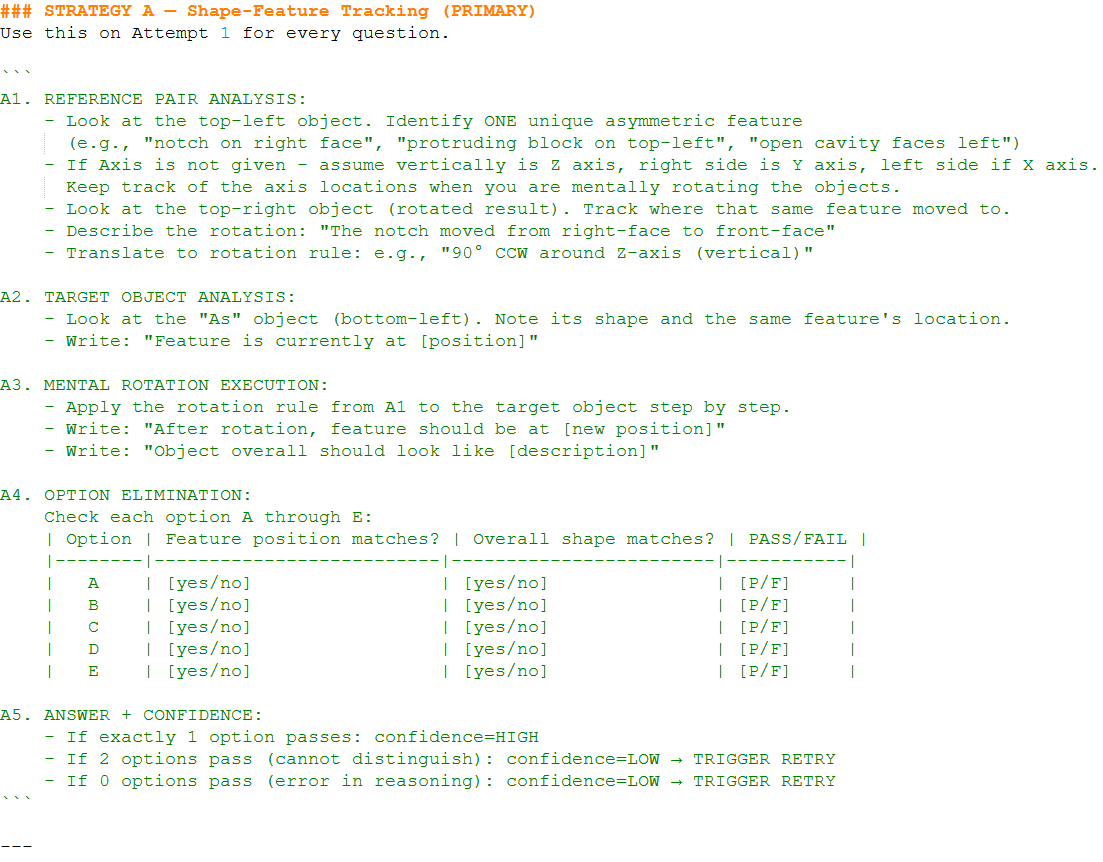}
    \caption{ Structured CoT Self-optimized prompting - Strategy A}
    \label{fig:A3}
\end{figure}

\begin{figure}[H]
    \centering
    \includegraphics[width=1\textwidth]{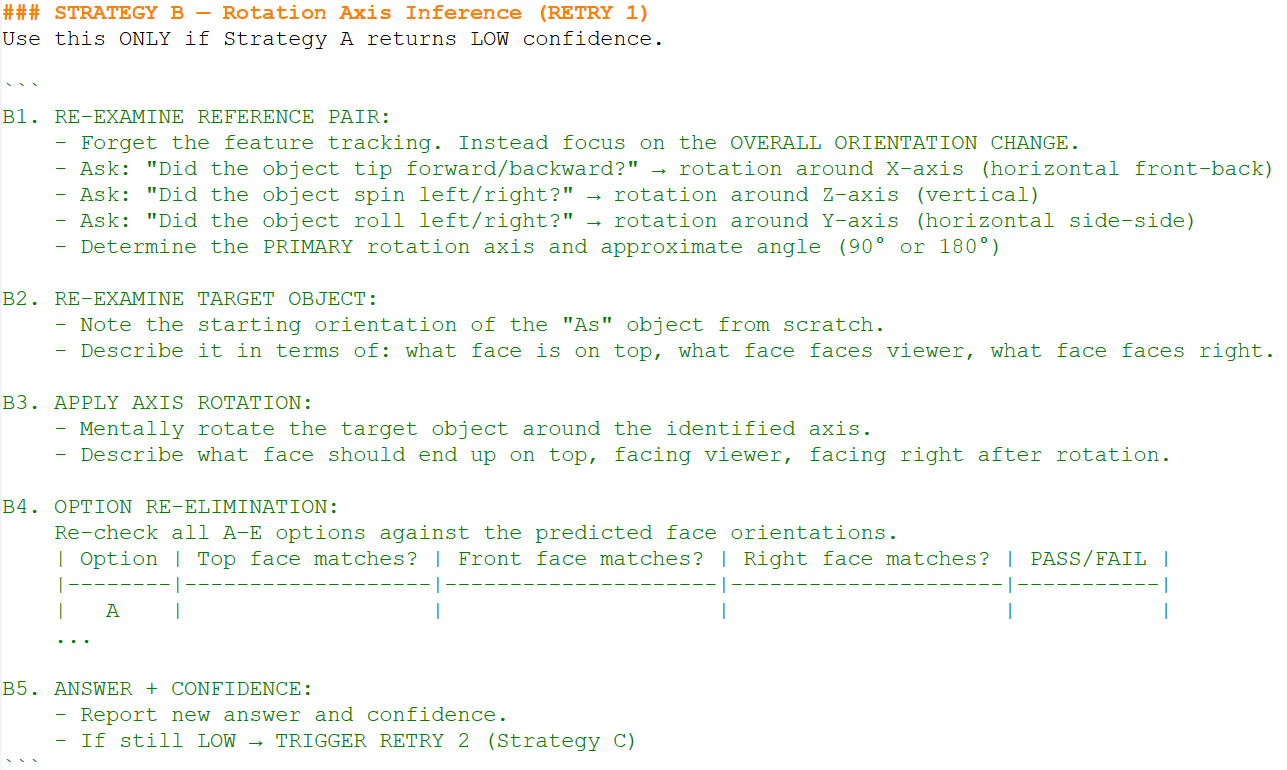}
    \caption{ Structured CoT Self-optimized prompting - Strategy B}
    \label{fig:A4}
\end{figure}

\begin{figure}[H]
    \centering
    \includegraphics[width=1\textwidth]{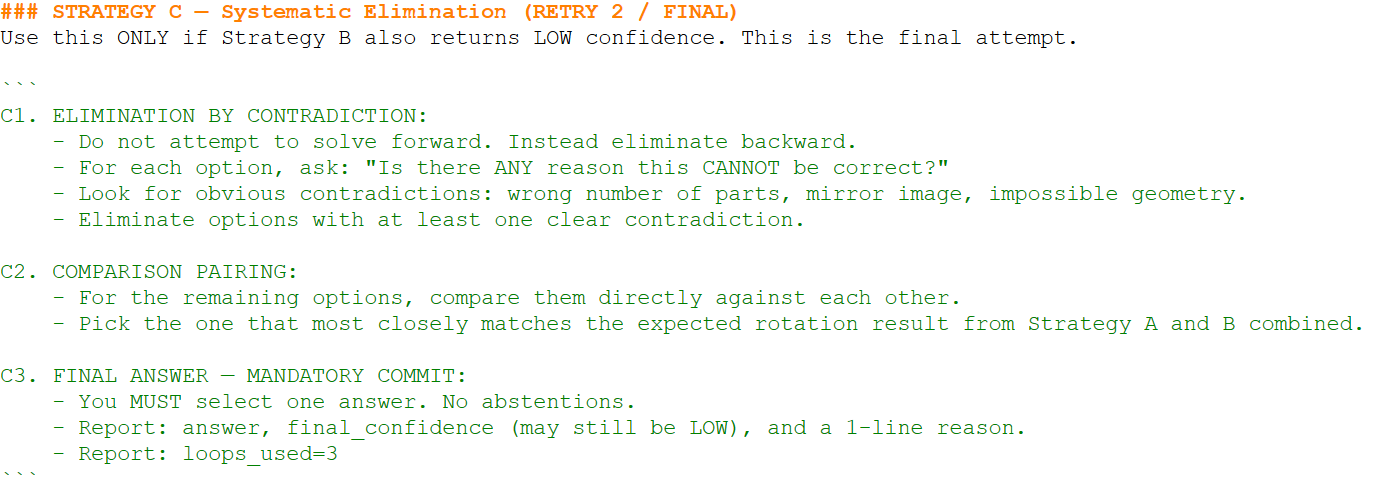}
    \caption{ Structured CoT Self-optimized prompting - Strategy C}
    \label{fig:A5}
\end{figure}

\begin{figure}[H]
    \centering

    \begin{subfigure}{\textwidth}
        \centering
        \includegraphics[width=1\textwidth]{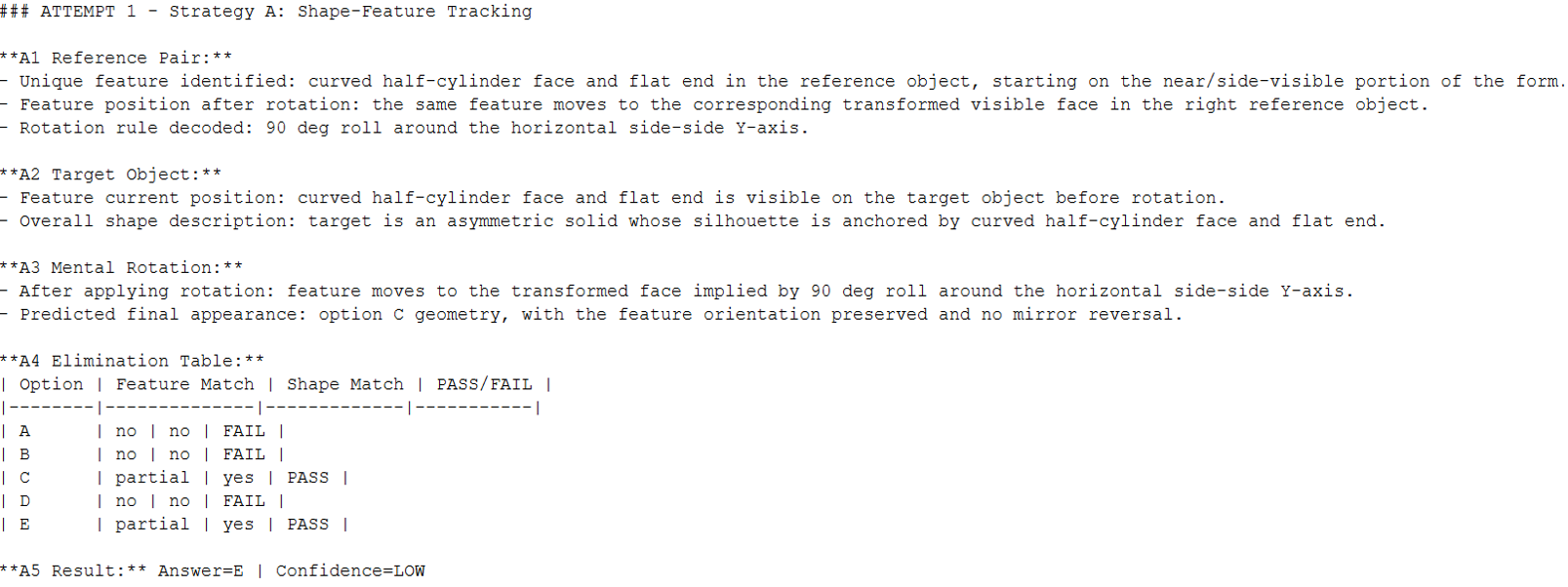}
        \caption{Sample answer self-optimized Structured CoT method (Strategy A)}
        \label{fig:rrr}
    \end{subfigure}

    \vspace{3pt}

    \begin{subfigure}{\textwidth}
        \centering
        \includegraphics[width=1\textwidth]{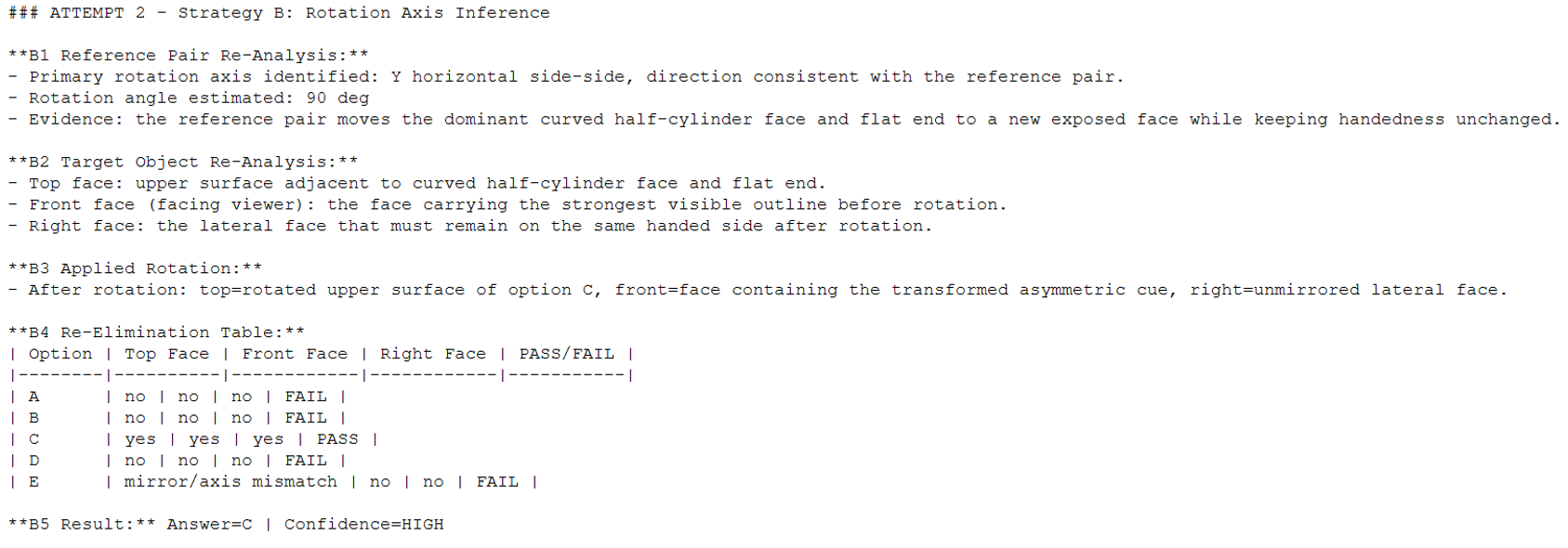}
        \caption{Sample answer self-optimized Structured CoT method (Strategy B)}
        \label{fig:your_label}
    \end{subfigure}
    
    \caption{Sample answer self-optimized Structured CoT method for the question of Revised PSVT:R (Correct answer C)}
    \label{fig:A6}
    
\end{figure}

\end{document}